\documentclass[10pt,english,oneside,twocolumn,a4paper]{article}
\usepackage{graphicx}
\usepackage[utf8]{inputenc}
\usepackage{times}
\fontfamily{ptm}\selectfont
\usepackage{tabularx}
\usepackage{ragged2e}
\usepackage[singlelinecheck=false]{caption}
\usepackage[T1]{fontenc}
\usepackage[numbers]{natbib}
\usepackage{amsmath}
\usepackage{amssymb}
\usepackage{booktabs}

\usepackage{babel}
\RequirePackage[blocks]{authblk}
\makeatletter
\let\ps@plain\ps@empty
\def\@xivpt{14pt}
\def\@sect#1#2#3#4#5#6[#7]#8{%
  \ifnum #2<2
    \null\par\vskip-15pt
  \fi
  \ifnum #2>\c@secnumdepth 
    \let\@svsec\@empty
  \else
    \refstepcounter{#1}%
    \protected@edef\@svsec{%
      \ifnum #2<4
        \hb@xt@10mm{\csname the#1\endcsname}\relax
      \else
        \hb@xt@12mm{\csname the#1\endcsname}\relax
      \fi}%
  \fi
  \@tempskipa #5\relax
  \ifdim \@tempskipa>\z@
    \begingroup
      #6{%
        \@hangfrom{\hskip #3\relax\@svsec}%
          \interlinepenalty \@M #8\@@par}%
    \endgroup
    \csname #1mark\endcsname{#7}%
    \addcontentsline{toc}{#1}{%
      \ifnum #2>\c@secnumdepth \else  
        \protect\numberline{\csname the#1\endcsname}%
      \fi 
      #7}%
  \else
    \def\@svsechd{%
      #6{\hskip #3\relax
      \@svsec #8}%
      \csname #1mark\endcsname{#7}%
      \addcontentsline{toc}{#1}{%
        \ifnum #2>\c@secnumdepth \else
          \protect\numberline{\csname the#1\endcsname}%
        \fi
        #7}}%
  \fi
  \@xsect{#5}}
\renewcommand\LARGE{\@setfontsize\LARGE{16}{20}}
\def\abstract#1{\def\@abstract{#1}}
\def\abstractEn#1{\def\@abstractEn{#1}}
\def\titleEn#1{\def\@titleEn{#1}}
\def\@maketitle{%
  \newpage
  \null
  \let \footnote \thanks
    {\LARGE\bfseries\RaggedRight \@title \par}%
    \vskip 1\baselineskip%
    {\normalsize
      \@author\par}%
    \vskip 2\baselineskip%
    \vskip \baselineskip%
    {\section*{Abstract}
      \@abstract}%
  \par
  \vskip 3\baselineskip}

\renewcommand\section{\@startsection {section}{1}{\z@}%
                                   {-3.5ex \@plus -1ex \@minus -.2ex}%
                                   {\baselineskip}%
                                   {\normalfont\Large\bfseries\RaggedRight}}
\renewcommand\subsection{\@startsection{subsection}{2}{\z@}%
                                     {\baselineskip}%
                                     {1ex}%
                                     {\normalfont\large\bfseries\RaggedRight}}
\renewcommand\subsubsection{\@startsection{subsubsection}{3}{\z@}%
                                     {1\baselineskip}%
                                     {3bp}%
                                     {\normalfont\normalsize\bfseries\RaggedRight}}
\renewcommand\paragraph{\@startsection{paragraph}{4}{\z@}%
                                    {1\baselineskip\@plus1ex \@minus.2ex}%
                                    {3bp}%
                                    {\normalfont\normalsize\RaggedRight}}
\renewcommand\subparagraph{\@startsection{subparagraph}{5}{\parindent}%
                                       {3.25ex \@plus1ex \@minus .2ex}%
                                       {-1em}%
                                      {\normalfont\normalsize\bfseries\RaggedRight}}
\parindent\p@
\parskip\z@
\DeclareCaptionLabelSeparator{enskip}{\enskip}
\makeatother

\title{OceanSAR-2: A Universal Feature Extractor for SAR Ocean Observation}

\author[a]{Alexandre Tuel}
\author[a]{Thomas Kerdreux}
\author[b]{Quentin Febvre}
\author[b]{Alexis Mouche}
\author[b]{Antoine Grouazel}
\author[c]{Jean-Renaud Miadana}
\author[a]{Antoine Audras}
\author[d]{Chen Wang}
\author[b]{Bertrand Chapron}
\affil[a]{Galeio, Paris, France}
\affil[b]{LOPS, Ifremer, Plouzané, France}
\affil[c]{OceanScope, Brest, France}
\affil[d]{Nanjing University of Information Science and Technology, Nanjing, China}

\abstract{
We present OceanSAR-2, the second generation of our foundation model for SAR-based ocean observation. Building on our earlier release, which pioneered self-supervised learning on Sentinel-1 Wave Mode data, OceanSAR-2 relies on improved SSL training and dynamic data curation strategies, which enhances performance while reducing training cost. OceanSAR-2 demonstrates strong transfer performance across downstream tasks, including geophysical pattern classification, ocean surface wind vector and significant wave height estimation, and iceberg detection. We release standardized benchmark datasets, providing a foundation for systematic evaluation and advancement of SAR models for ocean applications.
}

\begin{document}

\maketitle

\section{Introduction}

Recent advances in self-supervised learning (SSL) have profoundly changed how image representation models are developed and deployed. Instead of relying on large, task-specific labeled datasets, SSL enables learning directly from raw data, uncovering generic, transferable features that can serve as the backbone for a wide variety of downstream tasks. In computer vision, such so-called \textit{foundation models} have shown remarkable adaptability, reducing both annotation effort and the need to maintain multiple task-specific networks, not only for natural \cite{SiméoniDINOv32025} but also remote-sensing \cite{SzwarcmanPrithvi2025,JakubikTerraMind2025} imagery. In particular, this shift holds significant promise for Synthetic Aperture Radar (SAR), where labeled data are scarce and acquisition conditions vary widely.

For ocean remote sensing, in particular, the Sentinel-1 mission provides a unique source of global, high-resolution SAR data through its Wave Mode (WV) acquisitions. However, these data remain under-exploited due to the absence of large, well-curated labeled datasets and the difficulty of transferring models trained on land scenes or optical imagery. SSL offers a pathway toward universal SAR feature extractors capable of capturing rich and semantically meaningful ocean patterns, from swell and wind streaks to mesoscale structures and ice features, without requiring manual supervision.

In the broader field of remote sensing, the past few years have seen a surge of interest in building foundation models trained on large-scale satellite imagery datasets. Early works such as SatMAE \cite{CongSatMAE2025} and Prithvi \cite{SzwarcmanPrithvi2025} demonstrated that SSL on Earth observation data could yield representations transferable to many downstream applications. More recently, models such as TerraMind \cite{JakubikTerraMind2025}, Copernicus-FM \cite{WangCopernicus2025} and SkySenseV2 \cite{SkySense2025} have advanced this concept further through multi-modal learning, integrating optical, multispectral, SAR, and sometimes ancillary data such as elevation into shared embedding spaces.

Nevertheless, several limitations remain. First, most of these foundation models have been developed for land observation, emphasizing tasks such as land-cover classification or vegetation mapping. Very few have addressed ocean scenes, which differ substantially in both texture and semantics. Very few works have focused on Sentinel-1 Wave Mode imagery or, more broadly, on foundation models for ocean SAR observation. Second, while multi-modality offers richer representations, it typically requires larger and more complex architectures, making operational deployment more challenging. Moreover, multi-modal alignment assumes that optical and radar views describe the same physical content. By contrast, over the ocean, the SAR imaging mechanism is impacted by traveling waves during the synthetic aperture as well as different scattering mechanisms, leading to signatures that are rather different from what is visible in optical channels. This fundamental difference argues for maintaining SAR-specific representation models optimized for ocean physics. Third, most existing foundation models rely on carefully balanced datasets curated from global archives. This process requires significant manual effort and domain expertise. Such efforts have started to be made for RGB/multispectral imagery over land, with datasets such as SSL4EO-S12 \cite{wang2023ssl4eos12} or TerraMesh \cite{blumenstiel2025terramesh}. However, operational SAR ocean archives such as Sentinel-1 Wave Mode contain vast quantities of unevenly distributed and redundant data, which must be efficiently curated before large-scale SSL pretraining can be effective.

Based on these considerations, we previously introduced OceanSAR-1, a foundation model explicitly trained on ocean SAR imagery \cite{KerdreuxOceanSAR2025}. OceanSAR-1 was built upon the DINO self-supervised framework \cite{CaronDINO2018} and incorporated a novel dynamic data-pruning strategy designed to mitigate redundancy in the Sentinel-1 Wave Mode archive. This approach not only accelerated training convergence but also improved downstream performance by enhancing the diversity of effective training samples. The study further demonstrated that unimodal, domain-specific models can outperform more general multimodal foundation models for ocean observation tasks, while requiring significantly lower computational resources.

Building on these foundations, we introduce OceanSAR-2, a second-generation model that combines DINOv2-based pretraining with physically calibrated $\sigma^0$ inputs and an enhanced data curation pipeline. This combination substantially improves representation quality and downstream transfer while maintaining computational efficiency thanks to a relatively small model size. We also begin to lay the groundwork for the systematic and comprehensive benchmarking of foundation models for ocean SAR observation, an area where no standardized evaluation currently exists.

To this end, we propose a set of standardized evaluation datasets spanning key ocean applications, including pattern classification, geophysical parameter estimation, and object detection, and use them to assess OceanSAR-2's performance relative to prior models.

\section{Building OceanSAR-2}

\subsection{Self-supervised training strategy}

OceanSAR-2 builds directly on the design of OceanSAR-1 but introduces several improvements in both the training framework and input representation. While OceanSAR-1 relied on the original DINO SSL setup \cite{CaronDINO2018}, OceanSAR-2 adopts the more recent DINOv2 formulation \cite{OquabDINOv22024}, enabling the model to capture finer spatial structures. In both DINO and DINOv2, the self-supervised task aims to make a student network produce feature representations consistent with those of a teacher network when observing different parts (or "views") of the same image. The networks are also trained so that their embeddings remain invariant across different views of an image. This view-invariance objective drives the model to learn rich, semantically meaningful representations without explicit labels \cite{OquabDINOv22024}.

DINOv2 extends this principle with two key mechanisms. The first is a local-patch prediction loss (iBOT loss \cite{ZhouIBOT2022}) that encourages the model to align representations not only at the global image level but also across individual image patches. This improves sensitivity to small-scale structures and yields features that transfer better to segmentation or fine-grained detection tasks. The second is the KoLeo regularizer, which promotes a more uniform distribution of prototype vectors in feature space, thereby improving representation diversity and mitigating feature collapse.

\subsection{Training data and dataset curation}

Another important change from OceanSAR-1 concerns the input representation. Whereas OceanSAR-1 was trained directly on digital number (DN) amplitudes, OceanSAR-2 operates on calibrated $\sigma^0$ backscatter values. This physically meaningful normalization substantially improves feature consistency across acquisitions, incidence angles, and environmental conditions, making the learned representations more generalizable.

A persistent challenge in SAR pretraining is the lack of balanced datasets comparable to ImageNet. Balanced data is essential to optimise model generalisability and performance on downstream tasks. Sentinel-1 WV data are abundant but highly redundant in terms of concepts (pure ocean waves are much more common than cyclones or icebergs, for instance). Following the approach introduced with OceanSAR-1, we employ a dynamic pruning strategy that periodically selects the most diverse samples during training. This reduces over-representation of common patterns and promotes more balanced exposure to rarer phenomena such as icebergs or rain cells.

\subsection{Architecture}

\begin{figure}[h!]   
  \centering
  \includegraphics[width=69.5mm]{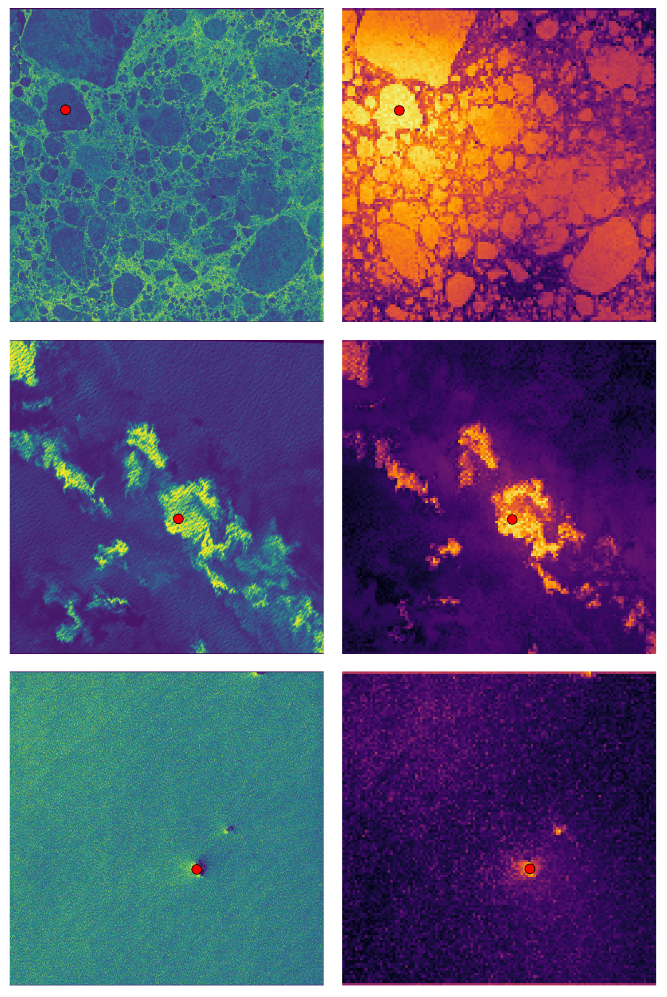}
  \caption{Example of Sentinel-1 WV images (left column; top: sea-ice, middle: rain cells; bottom: icebergs) with corresponding OceanSAR-2 feature similarity maps (right column). The reference patch to compute the similarity is indicated by a red dot in both panels.}\label{fig:1}
\end{figure}

The backbone architecture remains a Vision Transformer (ViT) that processes each image as a grid of patches ($16\times16$ pixels). For each patch, the network produces a $n$-dimensional embedding, and an additional class token of the same dimension summarizes the overall image content. A convenient way to inspect what the model has learned is to compute the cosine similarity between embeddings of a reference patch and all other patches in the same image. As illustrated in \textbf{Figure~\ref{fig:1}}, these feature-similarity maps reveal that OceanSAR-2 captures coherent and semantically meaningful structures: similar textures such as rain cells, sea ice, or icebergs are highlighted even when spatially distant.

\section{A benchmarking framework for SAR ocean observation}

\subsection{Current state}

Despite the growing importance of SAR data for ocean applications, no unified benchmarking framework currently exists for evaluating models across multiple ocean tasks. Existing studies typically focus on a single objective (\textit{e.g.}, wind retrieval, wave characterization, or iceberg detection) using purpose-built datasets and task-specific metrics. As a result, algorithmic advances remain difficult to compare or generalize. Moreover, a single dataset cannot capture the full spectrum of physical processes observable in SAR imagery: each phenomenon requires dedicated data sources, whether satellite or in-situ measurements from buoys or altimeters for geophysical parameters, or manually labelled data for object detection tasks. A partial exception is the ESA Sea State CCI initiative \cite{Plummer2017}, which has defined a benchmark for wave-height estimation, yet its scope is temporally limited and restricted to a single parameter. This situation highlights the need for a broader, evolving benchmarking philosophy suited to the dynamic nature of ocean missions and data streams.

A major challenge for ocean-SAR benchmarking lies in temporal and sensor continuity. Missions evolve over time, either through sensor replacement (\textit{e.g.}, Sentinel-1A to 1D) or changes in Level-0/1 processing chains, leading to systematic shifts in data characteristics. Static benchmarks, once established, therefore risk obsolescence. Instead, a "living benchmark" approach is needed: one that adapts to mission updates and data reprocessing while maintaining consistent evaluation protocols. Such a framework would ensure that model performance can be tracked meaningfully across generations of sensors and over the operational lifetime of Earth observation programs.

Designing a good benchmark also requires adherence to several fundamental principles. Accessibility is key: datasets should be publicly available online without restrictive access policies and rely on open-mission archives such as Sentinel-1. Each benchmark must have a clear and meaningful problem statement, grounded in scientific or operational relevance (for instance, estimating significant wave height or detecting icebergs). Reproducibility should be guaranteed through standard data formats, open-source reference code, and compatibility with mainstream data-science environments. Evaluation metrics must be analytically defined and transparent, enabling traceable comparisons across models. Finally, representativeness and longevity are essential: benchmarks should span diverse oceanic conditions (different wind regimes, latitudes, seasons, etc.) and be designed for regular updates as new data and sensors become available.

The benchmarks presented in this study aim to embody these principles. While currently limited to a subset of applications, they demonstrate how a modular, open, and evolving framework can be built for multi-application ocean-SAR evaluation. Such a foundation will be critical for future ESA and international efforts toward community-maintained, living benchmarks capable of assessing next-generation foundation models across sensors, modes, and ocean phenomena.

\subsection{Model benchmarking vs. fine-tuning}

When evaluating foundation models, it is useful to distinguish between \textit{representation benchmarking} and \textit{task optimization}. In the first case, benchmarks are used to assess the intrinsic quality of the learned representations. The goal is to compare the zero-shot performance of different model backbones, \textit{i.e.}, to measure how well their internal features transfer to new tasks without any retraining. In practice, most modern foundation models are based on Vision Transformers (ViTs), although some rely on convolutional backbones such as ResNet or ConvNeXt. These architectures produce, at various network depths, feature maps encoding spatial or semantic information. In addition, ViTs usually generate a single global embedding, often referred to as the class token (CLS), which summarizes the overall image content. Zero-shot regression or classification can then be performed directly from this embedding, for example by applying a k-nearest-neighbour (kNN) search or fitting a simple linear model on a small labeled subset. Such lightweight probing procedures provide a reproducible way to quantify the separability and semantic organization of the feature space, and are central to evaluating progress in self-supervised and representation learning research.

The second use of a benchmark is operational evaluation, where the aim is to achieve the highest possible performance on a well-defined application task which corresponds to the benchmark. In this setting, the benchmark acts as a standardized testbed for developing end-to-end solutions, which may include fine-tuning the backbone, designing specialized heads, or optimizing data augmentation and training regimes. Here, the absolute score matters more than isolating the contribution of the backbone itself. Both perspectives are valuable and complementary: representation benchmarking enables principled comparison and scientific understanding of model quality, while operational fine-tuning measures practical utility and identifies what performance levels are achievable in real-world conditions.

\subsection{Selected benchmarks}

We therefore propose a first version of our ocean workbench which consists of the four following datasets (Table \ref{table:1}. TenGeoP \cite{WangTenGeoP2019} and YOLOIB \cite{XieIcebergs2025} are manually labeled datasets and have previously been published; the other two, WV-SWH and WV-wind, rely on co-location with reference measurements and will be the object of an upcoming paper \cite{MoucheInPrep2025}.

\begin{table*}[t]
  \centering
  \caption{Overview of selected ocean-related benchmarks.}
  \label{table:1}
  \renewcommand{\arraystretch}{1.3} 
  \setlength{\tabcolsep}{6pt} 
  \begin{tabularx}{\textwidth}{|c|c|X|c|c|}
    \hline
    \textbf{Name} & \textbf{Task} & \textbf{Description} & \textbf{Labels} & \textbf{Reference} \\
    \hline
    TenGeoP & Classification & Classification of ocean scenes & Manual labels & \cite{WangTenGeoP2019} \\
    WV-SWH & Regression & Estimation of significant wave height & Altimeters & [in prep.] \\
    WV-wind & Regression & Estimation of surface wind speed and direction & HSCAT & [in prep.] \\
    YOLOIB & Object detection & Detection of icebergs in the southern hemisphere & Manual labels & \cite{XieIcebergs2025, WangYOLOIB2025} \\
    \hline
  \end{tabularx}
\end{table*}

\subsubsection{TenGeoP}

As a classification task, we selected the TenGeoP dataset \cite{WangTenGeoP2019}. It contains a total of 37,553 Sentinel-1 Wave Mode (WV) images that were manually classified by SAR specialists into ten distinct geophysical categories. These classes represent a range of oceanic and atmospheric phenomena, including pure ocean waves, wind streaks, micro-convective cells, rain cells, biological slicks, sea ice, icebergs, low-wind areas, as well as atmospheric and oceanic fronts. Each WV image is assigned a single label corresponding to its dominant pattern. The dataset provides global coverage and comprises exclusively Sentinel-1A acquisitions collected in 2016. For evaluation purposes, we reserved 20\% of the images (7,500 samples) as the test set, which we use here as benchmark.

\subsubsection{WV-SWH}

As a regression benchmark, we selected the estimation of Significant Wave Height (SWH). SWH is an operationally relevant parameter included in the ESA/Copernicus Sentinel-1 WV Level-2 ocean product and part of the ECVs (Essential Climate Variables) monitored by satellites through the ESA Climate Change Initiative (CCI) \cite{Plummer2017} program. SWH represents the mean height of the highest third of ocean waves passing a fixed point over a given period and is a key indicator of sea state conditions. To construct this dataset, we randomly selected 50,000 WV images across the global ocean and associated each image with a corresponding SWH value obtained from co-located altimeter measurements available in the CCI database \cite{Plummer2017}. A WV image and an altimeter record were considered co-located when acquired within a 3-hour time window and when the altimeter measurement lay within ±2° of the image center, with no flag activated. In cases where several altimeter observations satisfied these criteria, the spatially closest measurement was retained. A key element of using altimeter data as reference, instead of in-situ, is the number of colocations available on all ocean basins and spanning a wide range of sea states.  

\subsubsection{WV-wind}

As a second regression benchmark, we selected the estimation of ocean surface wind speed and direction. These two variables are also two geophysical parameters included in the ESA/Copernicus
Sentinel-1 WV Level-2 ocean product. We pair 50,000 WV images with co-located HSCAT wind measurements using stricter co-location criteria than for the WV-SWH dataset, since wind is typically less consistent in time and space than SWH. HSCAT swath width is also larger than that of altimeters, allowing for more true colocations. Specifically, we require that the HSCAT measurement has no rain flag activated, is within the SAR footprint and was taken within 30 minutes of the SAR image. In the rare case where several HSCAT measurements verify these conditions, we take their mean. As with the choice of altimeter for WV-SWH, we choose a remote sensing source here instead of in-situ to maximise the number of available colocations across ocean basins.

\subsubsection{YOLOIB}

Finally, we include a dense prediction task to our ocean workbench with the YOLOIB iceberg detection dataset introduced by \cite{XieIcebergs2025}. Icebergs were labeled by hand in 2,062 Sentinel-1A WV images collected during 2016.

\section{Results}

We illustrate the performance of OceanSAR2 on the ocean workbench introduced in the previous section, and compare it to that of two other existing foundation models trained on SAR imagery: a generalist one, TerraMind \cite{JakubikTerraMind2025} (whose training data included, among others, Sentinel-2 multispectral and Sentinel-1 SAR imagery, but neither ocean nor WV data), and a specialised model, WV-Net \cite{GlaserWVNet2025} (a SimCLR-based foundation model specifically dedicated to Sentinel-1 WV data, like the OceanSAR family of models). We also compare with the performance of the recently released DINOv3 model \cite{SiméoniDINOv32025}, trained on 3-channel (RGB), 30-cm satellite imagery exclusively. These models differ in terms of architecture and complexity. OceanSAR2 and WV-Net are small models ($\leq$ 25 million parameters), while TerraMind has 85 million and DINOv3 303 million parameters (distilled from a 7-billion parameter model \cite{SiméoniDINOv32025}). The dimension of the image embeddings produced by these models also varies: 384 for OceanSAR2, 768 for TerraMind, 1024 for DINOv3 and 2048 for WV-Net. 
Since TerraMind, unlike OceanSAR2 and WV-Net, does not provide a single, image-level embedding, but only feature maps, we apply a global max averaging to the final feature map (results are slightly better than with mean pooling).\\
We compare models both in zero-shot and fine-tuning mode. In zero-shot mode, we apply kNN classification and regression to the image features directly produced by the models (excluding the wind direction and iceberg benchmarks, for which this is not possible). As to fine-tuning, for the three regression tasks (wave height, wind speed and wind direction) we add a small, 3-layer MLP head (consisting of about 200,000-1,200,000 parameters, depending on the dimension of the model embeddings) which we fine-tune for 50 epochs with various learning rates ranging from 5E-5 to 5E-3, selecting the best one.
For the object detection task (icebergs), we add a classic DETR-like detection head \cite{carion2020end} with deformable attention \cite{zhudeformable} (from 6-12 million parameters, depending on the backbone), using the last feature map of each backbone as input, which we train for 500 epochs. We record the F1-score on predicted boxes, using an IoU threshold of 0.1 and score threshold of 0.5.

\begin{table}[ht!]
  \centering
  \resizebox{\columnwidth}{!}{
  \begin{tabular}{@{}lcccc@{}}
    \toprule
    Model & Params & TenGeoP (\%) & SWH (m) & Wspd (m/s) \\
    \midrule
    DINOv3 & 300M & \underline{91.9} & \underline{0.55} & \underline{1.68} \\
    TerraMind & 85M & 74.9 & 0.73 & 1.95 \\
    WV-Net & 24M & 91.5 & 0.64 & 1.71 \\
    \midrule
    OceanSAR2 & 21M & \textbf{94.0} & \textbf{0.52} & \textbf{1.32} \\
    \bottomrule
  \end{tabular}
  }
  \caption{Comparison of k-NN performance of existing models and ours across the ocean workbench. The number of parameters for each is indicated in the second column.}
  \label{table:1}
\end{table}

Results show that, in zero-shot mode, OceanSAR2 is the top performer on all benchmarks, with DINOv3 coming in second and WV-Net not far behind (Table \ref{table:2}). WV-Net is competitive on TenGeoP, with 91.5\% accuracy, but its performance worsens relative to OceanSAR2 on the other benchmarks (0.64 m vs. 0.52 m on WV-SWH and 1.71 m/s vs. 1.32 m/s on WV-wind). TerraMind, which was not trained on ocean SAR images, whether WV or else, has the worst performance, scoring 74.9\% on TenGeoP and about 10-20\% worse than WV-Net on the SWH and wind benchmarks.\\
Looking now at the MLP fine-tuning experiments, we see that OceanSAR2 still stands out with its top performance on 4 out of 5 benchmarks, but is equalled by DINOv3 (and WV-Net, almost) on TenGeoP, and surpassed by these two models on the WV-SWH dataset.

\begin{table}[ht!]
  \centering
  \resizebox{\columnwidth}{!}{
  \begin{tabular}{@{}lccccc@{}}
    \toprule
    Model & TenGeoP & SWH & Wspd & Wdir & YOLOIB \\
     & (\%) & (m) & (m/s) & ($^{\circ}$) & (F1@0.1)\\
    \midrule
    DINOv3 & \textbf{98.5} & \textbf{0.39} & \underline{1.12} & \underline{17.9} & * \\
    TerraMind & 87.1 & 0.67 & 1.82 & * & * \\
    WV-Net & \underline{98.3} & 0.427 & 1.23 & 21.4 & \underline{0.855} \\
    \midrule
    OceanSAR2 & \textbf{98.5} & \underline{0.40} & \textbf{1.01} & \textbf{16.9} & \textbf{0.865} \\
    \bottomrule
  \end{tabular}
  }
  \caption{Comparison of MLP fine-tuning performance of existing models and ours across the ocean workbench.\\
  (*: no reliable result: no available code or unstable fine-tuning). 
  }
  \label{table:2}
\end{table}

\section{Discussion and conclusion}

These results underscore the value of evaluating foundation models against multiple and heterogeneous benchmarks. A diverse set of tasks is essential to test whether a representation truly captures universal, semantically meaningful structures in ocean SAR imagery, rather than fitting narrowly to one phenomenon or metric. In this sense, TenGeoP serves as a useful first validation step but remains weakly discriminative: most models reach similarly high accuracy, close to the estimated label noise of 1–2\%.
This highlights that richer and more varied benchmarks are necessary to better differentiate representation quality.

Model-wise, several consistent trends emerge. DINOv3 delivers high performance, but at much higher computational cost for both inference and training (roughly three orders of magnitude more pretraining effort than OceanSAR-2), and with an order of magnitude more parameters ($\approx$300 M vs. 21 M). The version evaluated here was distilled from a multi-billion-parameter model, highlighting the computational inaccessibility of such approaches for most research and operational contexts. Conversely, OceanSAR-2, a compact, $\sigma^{0}$-native DINOv2 model, recovers much of this performance at a fraction of the cost, showing that architecture scale is not the sole determinant of representational power when the training data and objectives are carefully designed.

On SWH regression, WV-Net performs particularly well. This may be explained by its input normalization, which uses Sea Surface Roughness (SSR; $sigma^{0}$ corrected by a wind term) rather than $\sigma^{0}$ itself. SSR effectively emphasizes the normalized variance of the backscatter, a quantity known to be closely linked to wave modulation dynamics. By contrast, $\sigma^{0}$ directly reflects total backscatter power, which increases monotonically with wind speed. This difference in preprocessing likely gives WV-Net a physical advantage for wave-related regressions, underscoring that physics-informed data normalization remains an important complement to representation learning.

Finally, TerraMind underperforms across most tasks, even after fine-tuning. Beyond the lack of ocean-SAR data in its pretraining corpus, a likely cause is that its Masked Autoencoder (MAE) design favors dense reconstruction objectives. MAE features tend to be more spatially localized and less linearly separable, making them harder to leverage in classification or regression without deeper adaptation. While stronger task-specific fine-tuning could close the gap, our uniform fine-tuning setup was chosen precisely to compare backbones on equal footing. Under this constraint, models such as OceanSAR-2, whose objectives combine global invariance (DINO) with local consistency (iBOT) and feature diversity (KoLeo), offer a more transferable compromise.

Overall, these results demonstrate that a single, compact, $\sigma^{0}$-native backbone can act as a universal feature extractor for ocean SAR imagery, delivering strong performance across classification, regression, and detection tasks. This consistency suggests a clear operational pathway: centralized model maintenance with lightweight per-task adaptation, instead of multiple specialized models for each application.


This work confirms the potential of self-supervised learning to produce universal SAR feature extractors capable of representing complex oceanic patterns without manual supervision. We showed that OceanSAR-2, trained efficiently on Sentinel-1 Wave Mode data, achieves competitive results across diverse benchmarks despite its compact size and modest training cost. Beyond academic validation, this offers a practical foundation for operational pipelines that require robustness, interpretability, and ease of maintenance.

Broader benchmark coverage will be key to sustaining progress. Datasets such as TenGeoP provide a valuable starting point, but future benchmarking should encompass a wider range of operational problems: ship detection (AIS), oil slicks (EMSA), internal waves, MABL parameters, and more. Complementary data sources, including in-situ networks (drifting or fixed buoys), altimetry, SWOT, and passive microwave sensors, can further extend the realism and discriminative power of such evaluations. Establishing this ecosystem of datasets will enable more robust comparisons and accelerate the maturation of ocean-SAR foundation models.

Looking ahead, OceanSAR-2 provides a strong basis for expanding to new sensing regimes. The same training and curation principles can be applied to other SAR imaging modes (e.g., IW or EW), enabling analysis of cross-resolution and cross-geometry generalization. Extending to multi-polarization and even phase-sensitive inputs would open new possibilities for modeling ocean roughness, atmospheric stability, or sea ice/iceberg dynamics in a unified representation space. Finally, cross-mission pretraining that spans multiple sensors and radar bands, whether C or L band, offers a clear path toward sensor-agnostic ocean-SAR foundation models, capable of integrating heterogeneous archives into a coherent representation framework.

\bibliographystyle{IEEEtranN}
\bibliography{biblio}

@misc{WangYOLOIB2025,
  title={{YOLOIB code and data}},
  author={Xie, Tao and Wang, Chen and Li, Xiao-Ming and Li, Hui-Min and Fan, Gaojing and Mouche, Alexis and Chapron, Bertrand},
  url={https://zenodo.org/records/17216910},
  year={2025}
}

@article{MoucheInPrep2025,
  title={{A benchmarking framework for ocean SAR observation}},
  author={Mouche, Alexis and Miadana, Jean-Renaud and Grouazel, Antoine and Tuel, Alexandre and Kerdreux, Thomas and Febvre, Quentin and Chapron, Bertrand},
  journal={in prep.},
  year={2026}
}

@article{XieIcebergs2025,
  title={{Iceberg Detection from the Global Sentinel-1 Wave Mode SAR Data with YOLO Deep Learning}},
  author={Xie, Tao and Wang, Chen and Li, Xiao-Ming and Li, Hui-Min and Fan, Gaojing and Mouche, Alexis and Chapron, Bertrand},
  journal={SSRN},
  doi={10.2139/ssrn.5548767},
  year={2025}
}

@article{SkySense2025,
  title={{A semantic-enhanced multi-modal remote sensing foundation model for Earth observation}},
  author={Wu, Kang and Zhang, Yingying and Ru, Lixiang and Dang, Bo and Lao, Jiangwei and Yu, Lei and Luo, Junwei and Zhu, Zifan and Sun, Yue and Zhang, Jiahao and Zhu, Qi and Wang, Jian and Yang, Ming and Chen, Jingdong and Zhang, Yongjun and Li, Yansheng},
  journal={Nature Machine Intelligence},
  pages={1235-1249},
  volume={7},
  issue={8},
  doi={10.1038/s42256-025-01078-8},
  year={2025}
}

@misc{wang2023ssl4eos12,
      title={{SSL4EO-S12: A Large-Scale Multi-Modal, Multi-Temporal Dataset for Self-Supervised Learning in Earth Observation}}, 
      author={Yi Wang and Nassim Ait Ali Braham and Zhitong Xiong and Chenying Liu and Conrad M Albrecht and Xiao Xiang Zhu},
      year={2023},
      eprint={2211.07044},
      archivePrefix={arXiv},
}

@article{Plummer2017,
    title = {{The ESA Climate Change Initiative (CCI): A European contribution to the generation of the Global Climate Observing System}},
    journal = {Remote Sensing of Environment},
    volume = {203},
    pages = {2-8},
    year = {2017},
    note = {Earth Observation of Essential Climate Variables},
    issn = {0034-4257},
    doi = {10.1016/j.rse.2017.07.014},
    author = {Stephen Plummer and Pascal Lecomte and Mark Doherty},
}

@misc{blumenstiel2025terramesh,
      title={{TerraMesh: A Planetary Mosaic of Multimodal Earth Observation Data}}, 
      author={Benedikt Blumenstiel and Paolo Fraccaro and Valerio Marsocci and Johannes Jakubik and Stefano Maurogiovanni and Mikolaj Czerkawski and Rocco Sedona and Gabriele Cavallaro and Thomas Brunschwiler and Juan Bernabe-Moreno and Nicolas Longépé},
      year={2025},
      eprint={2504.11172},
      archivePrefix={arXiv},
}

@misc{SzwarcmanPrithvi2025,
      title={{Prithvi-EO-2.0: A Versatile Multi-Temporal Foundation Model for Earth Observation Applications}}, 
      author={Daniela Szwarcman and Sujit Roy and Paolo Fraccaro and Þorsteinn Elí Gíslason and Benedikt Blumenstiel and Rinki Ghosal and Pedro Henrique de Oliveira and Joao Lucas de Sousa Almeida and Rocco Sedona and Yanghui Kang and Srija Chakraborty and Sizhe Wang and Carlos Gomes and Ankur Kumar and Myscon Truong and Denys Godwin and Hyunho Lee and Chia-Yu Hsu and Ata Akbari Asanjan and Besart Mujeci and Disha Shidham and Trevor Keenan and Paulo Arevalo and Wenwen Li and Hamed Alemohammad and Pontus Olofsson and Christopher Hain and Robert Kennedy and Bianca Zadrozny and David Bell and Gabriele Cavallaro and Campbell Watson and Manil Maskey and Rahul Ramachandran and Juan Bernabe Moreno},
      year={2025},
      eprint={2412.02732},
      archivePrefix={arXiv},
}

@misc{WangCopernicus2025,
      title={{Towards a Unified Copernicus Foundation Model for Earth Vision}}, 
      author={Yi Wang and Zhitong Xiong and Chenying Liu and Adam J. Stewart and Thomas Dujardin and Nikolaos Ioannis Bountos and Angelos Zavras and Franziska Gerken and Ioannis Papoutsis and Laura Leal-Taixé and Xiao Xiang Zhu},
      year={2025},
      eprint={2503.11849},
      archivePrefix={arXiv},
}

@InProceedings{KerdreuxOceanSAR2025,
    author    = {Kerdreux, Thomas and Tuel, Alexandre and Febvre, Quentin and Mouche, Alexis and Chapron, Bertrand},
    title = {{Efficient Self-Supervised Learning for Earth Observation via Dynamic Dataset Curation}},
    booktitle = {Proceedings of the IEEE/CVF Conference on Computer Vision and Pattern Recognition (CVPR) Workshops},
    year = {2025},
    pages = {3017-3027}
}

@InProceedings{CaronDINO2018,
  author={Caron, Mathilde and Touvron, Hugo and Misra, Ishan and Jegou, Hervé and Mairal, Julien and Bojanowski, Piotr and Joulin, Armand},
  booktitle={2021 IEEE/CVF International Conference on Computer Vision (ICCV)}, 
  title={{Emerging Properties in Self-Supervised Vision Transformers}}, 
  year={2021},
  pages={9630-9640},
  doi={10.1109/ICCV48922.2021.00951}
}

@article {GlaserWVNet2025,
      author = "Yannik Glaser and Justin E. Stopa and Linnea M. Wolniewicz and Ralph Foster and Doug Vandemark and Alexis Mouche and Bertrand Chapron and Peter Sadowski",
      title = {{WV-Net: A Foundation Model for SAR Ocean Satellite Imagery}},
      journal = "Artificial Intelligence for the Earth Systems",
      year = "2025",
      volume = "4",
      number = "4",
      doi = "10.1175/AIES-D-25-0003.1",
      pages = "250003",
}

@inproceedings{CongSatMAE2025,
 author = {Cong, Yezhen and Khanna, Samar and Meng, Chenlin and Liu, Patrick and Rozi, Erik and He, Yutong and Burke, Marshall and Lobell, David and Ermon, Stefano},
 booktitle = {Advances in Neural Information Processing Systems},
 editor = {S. Koyejo and S. Mohamed and A. Agarwal and D. Belgrave and K. Cho and A. Oh},
 pages = {197--211},
 publisher = {Curran Associates, Inc.},
 title = {{SatMAE: Pre-training Transformers for Temporal and Multi-Spectral Satellite Imagery}},
 volume = {35},
 year = {2022}
}

@misc{JakubikTerraMind2025,
      title={{TerraMind: Large-Scale Generative Multimodality for Earth Observation}}, 
      author={Johannes Jakubik and Felix Yang and Benedikt Blumenstiel and Erik Scheurer and Rocco Sedona and Stefano Maurogiovanni and Jente Bosmans and Nikolaos Dionelis and Valerio Marsocci and Niklas Kopp and Rahul Ramachandran and Paolo Fraccaro and Thomas Brunschwiler and Gabriele Cavallaro and Juan Bernabe-Moreno and Nicolas Longépé},
      year={2025},
      eprint={2504.11171},
      archivePrefix={arXiv},
}

@misc{OquabDINOv22024,
      title={{DINOv2: Learning Robust Visual Features without Supervision}}, 
      author={Maxime Oquab and Timothée Darcet and Théo Moutakanni and Huy Vo and Marc Szafraniec and Vasil Khalidov and Pierre Fernandez and Daniel Haziza and Francisco Massa and Alaaeldin El-Nouby and Mahmoud Assran and Nicolas Ballas and Wojciech Galuba and Russell Howes and Po-Yao Huang and Shang-Wen Li and Ishan Misra and Michael Rabbat and Vasu Sharma and Gabriel Synnaeve and Hu Xu and Hervé Jegou and Julien Mairal and Patrick Labatut and Armand Joulin and Piotr Bojanowski},
      year={2024},
      eprint={2304.07193},
      archivePrefix={arXiv},
}

@misc{ZhouIBOT2022,
      title={{iBOT: Image BERT Pre-Training with Online Tokenizer}}, 
      author={Jinghao Zhou and Chen Wei and Huiyu Wang and Wei Shen and Cihang Xie and Alan Yuille and Tao Kong},
      year={2022},
      eprint={2111.07832},
      archivePrefix={arXiv},
}

@misc{SiméoniDINOv32025,
      title={{DINOv3}}, 
      author={Oriane Siméoni and Huy V. Vo and Maximilian Seitzer and Federico Baldassarre and Maxime Oquab and Cijo Jose and Vasil Khalidov and Marc Szafraniec and Seungeun Yi and Michaël Ramamonjisoa and Francisco Massa and Daniel Haziza and Luca Wehrstedt and Jianyuan Wang and Timothée Darcet and Théo Moutakanni and Leonel Sentana and Claire Roberts and Andrea Vedaldi and Jamie Tolan and John Brandt and Camille Couprie and Julien Mairal and Hervé Jégou and Patrick Labatut and Piotr Bojanowski},
      year={2025},
      eprint={2508.10104},
      archivePrefix={arXiv},
}

@article{WangTenGeoP2019,
title = {{Classification of the global Sentinel-1 SAR vignettes for ocean surface process studies}},
journal = {Remote Sensing of Environment},
volume = {234},
pages = {111457},
year = {2019},
issn = {0034-4257},
doi = {10.1016/j.rse.2019.111457},
author = {Chen Wang and Pierre Tandeo and Alexis Mouche and Justin E. Stopa and Victor Gressani and Nicolas Longepe and Douglas Vandemark and Ralph C. Foster and Bertrand Chapron},
}

@inproceedings{zhudeformable,
  title={{Deformable DETR: Deformable Transformers for End-to-End Object Detection}},
  author={Zhu, Xizhou and Su, Weijie and Lu, Lewei and Li, Bin and Wang, Xiaogang and Dai, Jifeng},
  booktitle={International Conference on Learning Representations},
  year={2021}
}

@inproceedings{carion2020end,
  title={End-to-end object detection with transformers},
  author={Carion, Nicolas and Massa, Francisco and Synnaeve, Gabriel and Usunier, Nicolas and Kirillov, Alexander and Zagoruyko, Sergey},
  booktitle={European conference on computer vision},
  pages={213--229},
  year={2020},
  organization={Springer}
}

\end{document}